\documentclass{article}
\usepackage{ijcai25}

\usepackage{times}
\usepackage{soul}
\usepackage{url}
\usepackage[hidelinks]{hyperref}
\usepackage[utf8]{inputenc}
\usepackage[small]{caption}
\usepackage{graphicx}
\usepackage{amsmath}
\usepackage{amsthm}
\usepackage{booktabs}
\usepackage{algorithm}
\usepackage{algorithmic}
\usepackage[switch]{lineno}
\usepackage{float}
\usepackage{placeins}

\nolinenumbers

\newcommand{\correspondingauthor}{\textsuperscript{*}}

\title{An Atomic Skill Library Construction Method for Data-Efficient Embodied Manipulation}

\author{
Dongjiang Li$^1$
\and
Bo Peng$^{1,3}$\and
Chang Li$^1$\and
Ning Qiao$^2$\and
Qi Zheng$^4\correspondingauthor$\and
Lei Sun$^5$\and
Yusen Qin$^{6,7}$\and
Bangguo Li$^6$\and
Yifeng Luan$^2$\and
Bo Wu$^8$\and
Yibing Zhan$^1$\and
Mingang Sun$^{2,6}$\and
Tong Xu$^3\correspondingauthor$\and
Lusong Li$^1$\and
Hui Shen$^2$\And
Xiaodong He$^1$\\
\affiliations
$^1$JD Explore Academy, China\\
$^2$Jingdong Technology Information Technology Co., Ltd\\
$^3$University of Science and Technology of China\\
$^4$Shenzhen University \\
$^5$Haier Group\\
$^6$Tsinghua University\\
$^7$D-robotics\\
$^8$RealMan Intelligent  Technology (Jiangsu/Beijing) Co., Ltd\\
\emails
\{lidongjiang5, pengbo104, lichang93, qiaoning7, luanyifeng, sunmingang, shenhui, hexiaodong\}@jd.com,
\{ssdzxq, leeakalm2002, lilusong\}@gmail.com,
sunlei.uh@haier.com,
yusen.qin@d-robotics.cc,
bo@realman-robot.com,
zybjy@mail.ustc.edu.cn,
tongxu@ustc.edu.cn
}

\begin{document}

\maketitle

\begin{abstract}
Embodied manipulation is a fundamental ability in the realm of embodied artificial intelligence. Although current embodied manipulation models show certain generalizations in specific settings, they struggle in new environments and tasks due to the complexity and diversity of real-world scenarios. The traditional end-to-end data collection and training manner leads to significant data demands. Decomposing end-to-end tasks into atomic skills helps reduce data requirements and improves the task success rate. However, existing methods are limited by predefined skill sets that cannot be dynamically updated. To address the issue, we introduce a three-wheeled data-driven method to build an atomic skill library. We divide tasks into subtasks using the Vision-Language-Planning (VLP). Then, atomic skill definitions are formed by abstracting the subtasks. Finally, an atomic skill library is constructed via data collection and Vision-Language-Action (VLA) fine-tuning. As the atomic skill library expands dynamically with the three-wheel update strategy, the range of tasks it can cover grows naturally. In this way, our method shifts focus from end-to-end tasks to atomic skills, significantly reducing data costs while maintaining high performance and enabling efficient adaptation to new tasks. Extensive experiments in real-world settings demonstrate the effectiveness and efficiency of our approach.
\end{abstract}

\let\thefootnote\relax\footnotetext{\correspondingauthor Corresponding authors.}

\section{Introduction}
Embodied intelligence, primarily referring to “embodied artificial intelligence”, has seen significant advances in the era of generative AI. By mapping multimodal data, such as text, images, and speech into a unified semantic continuous vector space, this enables domain-independent cross-modal integration. This space closely links the semantic discrete symbol space with the feature continuous vector space, providing a new opportunity for embodied intelligence to develop into a general form. End-to-end embodied manipulation, especially Vision-Language-Action (VLA) models, have shown significant progress due to the availability of embodied data \cite{o2023open} \cite{khazatsky2024droid} and the advancement of multimodal technologies, demonstrating increasing generality and generalization, greatly enhancing the potential for service robots to be practically applied.



Although current embodied manipulation models demonstrate a certain generalization in specific settings, they face challenges in adapting to new environments and tasks \cite{black2024pi_0} \cite{li2024cogact}. Notably, current embodied manipulation models are characterized by their end-to-end orientation, where both data collection and model training are performed based on specific end-to-end tasks. In scientific research, a certain number of end-to-end tasks can be manually defined as standards to facilitate the improvement of algorithm performance \cite{team2024octo} \cite{liu2024rdt}. However, the diversity and complexity of real-world scenarios make the end-to-end method infeasible for the practical application in general embodied manipulation. On the one hand, real-world tasks are impossible to enumerate, while the end-to-end method fails to extend the learned task abilities to new tasks. On the other hand, as the complexity of a task increases, such as with more complicated procedures, the data requirements of the task must increase to maintain satisfactory performance. These problems make every single task require a significant number of corresponding new data~\cite{kim2024openvla}~\cite{liu2024rdt}, leading to the risk of ``data explosion", which significantly hinders the practical application of embodied manipulation models. 

Given these challenges associated with end-to-end methods, a natural solution is to break down the execution of an end-to-end task into the sequential execution of atomic skills \cite{diab2020skillman} \cite{mao2024robomatrix}. These atomic skills are more fine-grained and general, allowing reuse across different scenarios and tasks. Furthermore, by our experiments, we found that mapping end-to-end tasks\cite{team2024octo} \cite{kim2024openvla} into atomic skills helps reduce the data requirement and improve the task execution success rate. Although previous work on atomic skills has demonstrated effectiveness, it is limited by the manual predefinition of specific skills. The predefined skill sets are highly limited and incapable of dynamic updating and expansion, thus remaining highly inefficient for general embodied manipulation in the real world. Actually, automatic task decomposition, planning and skill definition are feasible based on today's powerful unimodal and multimodal understanding and reasoning technologies. 

In light of the potential of atomic skills and the limitations of prior arts, our work focuses on developing a practical atomic skill library construction method to achieve data-efficient embodied manipulation in real-world scenarios. Specifically, the proposed method consists of three main modules, termed as ``three-wheeled'' method because each module can update as the development of relevant technology. Specifically, during training stage, the Vision-Language-Planning (VLP) module, which integrates visual perception, language understanding, and spatial intelligence, is employed to decompose the given task into multiple subtasks. These subtasks are mapped into a set of atomic skill definitions by the high-level semantic abstraction module based on the plasticity and adaptability of the backbone VLA. Corresponding data collection is performed oriented towards the automatically generated atomic skill definitions, which are used for few-shot learning of the VLA module to achieve skill library. Compared to traditional end-to-end strategies, our method achieves the exponential improvement of data collection efficiency and realizes the cross-task and cross-scenario generality. The contributions of this paper are as follows:

\begin{itemize}
\item We propose a novel three-wheeled framework that combines VLP, responsible for task decomposition and planning, with VLA, responsible for skill execution, to construct an atomic skill library in a data-driven manner.
\item Based on this framework, we implement a VLP agent capable of effective task decomposition and real-time planning, along with a semantic abstraction strategy that maps subtasks to a set of general atomic skill definitions, integrated with fine-tuning VLA to realize a practical method for atomic skill library construction.
\item To the best of our knowledge, we are among the first to attempt to address the data explosion in practical applications of embodied manipulation methods through an atomic skill library approach. Extensive experiments in real-world settings show that our atomic skill library method can significantly reduce data costs while demonstrating excellent task execution capabilities.
\end{itemize}


\begin{figure*}[t]
	\centering
	\includegraphics[width=1.0\textwidth]{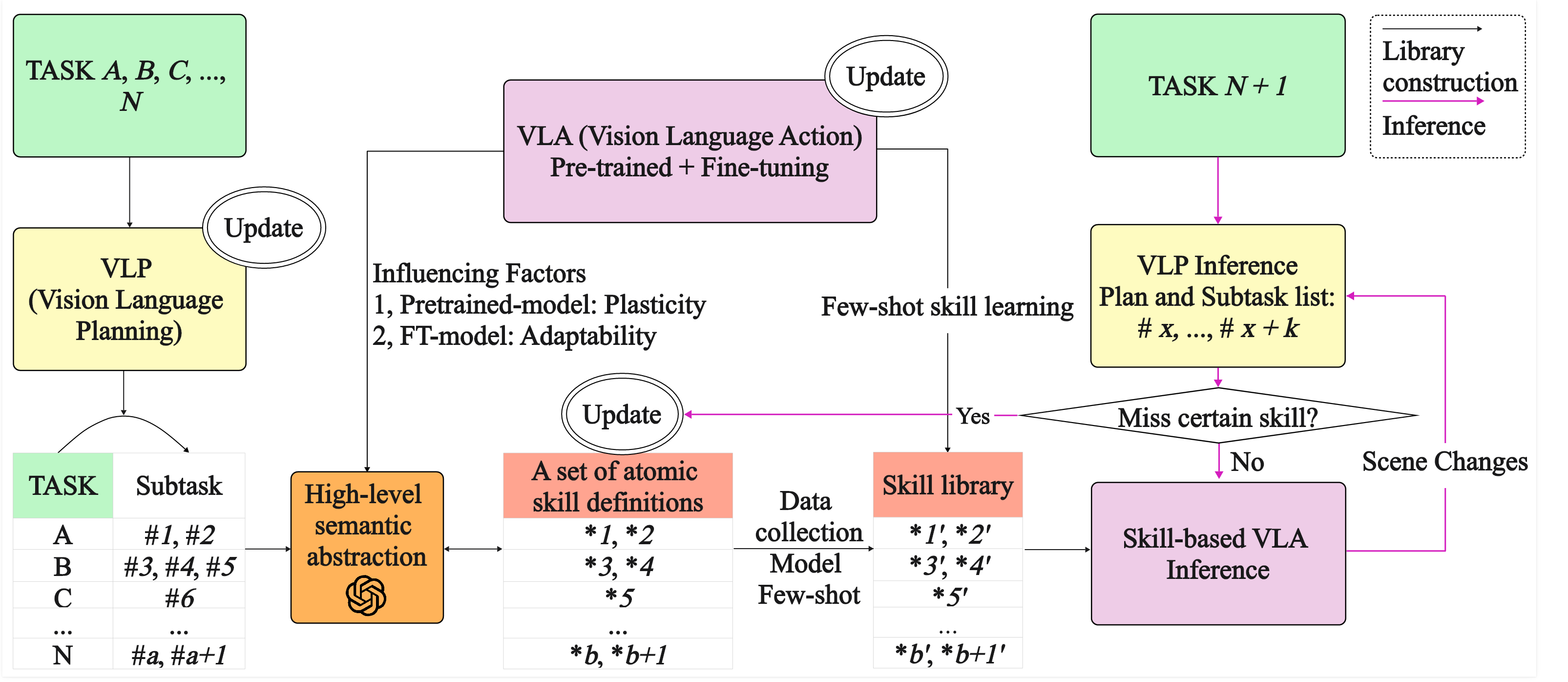}
	\caption{Three-Wheeled Self-Driven Atomic Skill Library Construction and Inference Pipeline.}
	\label{fig:framework}	
\end{figure*}

\section{Related work}
\paragraph{Vision-Language-Action Models:} Recent advances in VLA models demonstrate significant potential to enable end-to-end embodied manipulation tasks. These models typically employ an end-to-end training paradigm that integrates visual, language, and action data to understand and execute task instructions. Such multimodal integration empowers robots to handle sophisticated tasks in dynamic and unstructured environments. 


OpenVLA \cite{kim2024openvla} achieves effective robot control through parameter-efficient fine-tuning and quantization tailored to new target scenarios. Meanwhile, diffusion models \cite{ho2020denoising} \cite{song2020denoising} \cite{rombach2022high} have emerged as the backbone networks for numerous applications, owing to their exceptional expressiveness and high-quality sampling capabilities. In robotics, diffusion models have become a preferred choice for policy representation. Diffusion Policy \cite{chi2023diffusion}, for instance, pioneers the use of conditional denoising diffusion processes to model robotic behaviors, effectively representing visual motion strategies. Octo \cite{team2024octo} is a large-scale, general-purpose policy, and supports diverse task guidance through either language commands or target images. RDT-1B \cite{liu2024rdt} leverages Diffusion Transformers (DiTs) \cite{peebles2023scalable} as its scalable backbone, delivering remarkable performance on dual-arm robotic platforms. Recently, researchers \cite{black2024pi_0} have proposed a flow-matching action chunking architecture \cite{zhao2023learning} to model complex continuous action distributions. 

Although VLA models have made significant progress in robotic control, current embodied manipulation approaches still face critical challenges, particularly the need for massive amounts of data to learn and generalize complex tasks effectively. Despite these challenges, our work explores the decomposition of complex tasks into smaller atomic tasks, enabling robots to learn the corresponding atomic skills for each.

\paragraph{Vision-Language-Planning Models:} 
VLP refers to task planning that combines visual information and language instruction in embodied manipulation. When facing real-world tasks, direct end-to-end execution is challenging and often results in poor performance. This deficiency makes effective task decomposition and planning in urgent demand.

UOF \cite{yang2021hierarchical} incorporates planning and control of intermediate steps in complex tasks, enabling robots to learn different outcomes in multi-step tasks. ECoT \cite{zawalski2024robotic} has introduced the Chain-of-Thought (CoT) approach, in which they train VLAs to perform multiple steps of reasoning about plans, sub-tasks, motions, and visually grounded features like object bounding boxes and end effector positions, before predicting the robot action. RoboMatrix\cite{mao2024robomatrix} directly uses proprietary VLMs to perform task decomposition and existing object detection models to check execution status based on the judgment of object existence.

Although previous VLP methods show certain effectiveness, they are limited to basic visual information like objects and scene description while lacking spatial intelligence, which is vital to real-world embodied manipulation.

\paragraph{Atomic Skills:} 
Due to the complexity of tasks in real-world scenarios, some researchers have attempted to introduce the concept of atomic skills to decompose tasks. The SkillMaN \cite{diab2020skillman} framework includes a module with experiential knowledge that demonstrates how to execute a set of skills using workflows and robotic trajectories. For complex tasks, some researchers \cite{kroemer2021review} focus on the hierarchical decomposition of tasks and the reusability of robotic skills. Recent approaches have incorporated task planning by using an agent to automatically break down complex tasks into atomic tasks, rather than relying on manual decomposition. For instance, researchers \cite{zhao2022system} have used the concepts of primitives, skill decomposition, and synthesis to analyze the operational skills involved in robot bomb disposal tasks, and proposed a knowledge-based approach for learning these operational skills. RoboMatrix \cite{mao2024robomatrix} introduced the concept of meta-skills, while limited to a manual pre-defined skill set. It does not effectively address data explosion and cannot support the update of VLP and VLA modules.

\section{Method}
\label{method}
\subsection{Overview}
Our goal is to develop a data-driven approach for generating a skill library, enabling continuous self-updating through three integrated sub-modules. The framework of the proposed method is illustrated in Fig.~\ref{fig:framework}. Specifically, for task instructions provided by users, the VLP module breaks down the instructions into corresponding subtasks. Next, a high-level semantic abstraction module abstracts these subtasks into a set of general atomic skill definitions. Finally, we collect data and fine-tune VLA models to acquire atomic skills, ultimately constructing a comprehensive skill library. 

When faced with a new task, our framework calls VLP for task planning and retrieves the corresponding atomic skills from the skill library. If all the required skills are covered by the library, the task will be executed without any additional data collecting and fine-tuning. If a specific skill is missing, the high-level semantic abstraction module will wake up to update the set of atomic skills, and only the trajectory data of the missing skill are needed. For example, if a new task ``Give the guest a cup of water'' is proposed and the current atomic skill library includes the skills ``Lift up the bottle'' and ``Align and tilt the bottle towards the cup'', only a new skill ``deliver the cup'' is needed. In this way, our method enables the execution of a new task with little or even no extra data requirement, effectively enhancing the data efficiency and generality.

\begin{figure*}[t]
\centering
\includegraphics[width=1.0\textwidth]{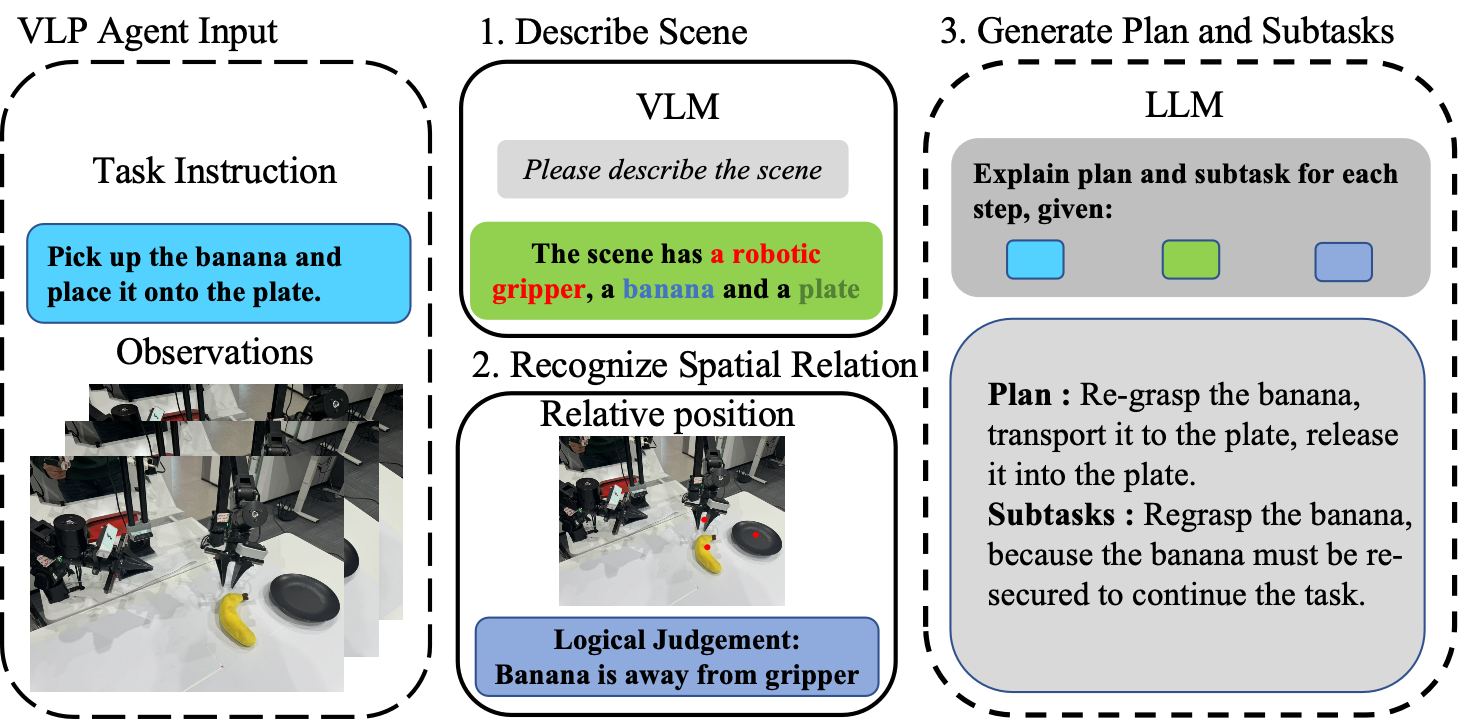}
\caption{The VLP agent reasoning chain framework based on spatial intelligence information.}
\label{fig:VLP}
\end{figure*}

\subsection{VLP Agent Wheel}
To uniformly achieve task decomposition during training and task planning during inference, we build a VLP agent that integrates visual perception, language understanding, and spatial intelligence.

As shown in Fig.~\ref{fig:VLP}, given a text-based task instruction and image-based current observation, the Prismatic model \cite{karamcheti2024prismatic}, an off-shelf VLM, is introduced to generate a scene description corresponding to the observation image. Considering the spatial complexity of the 3D world, we additionally devise a strategy for spatial relationship awareness. We first use Dino-X~\cite{ren2024dino}, a remarkable object detection model, to detect task-related objects in the observation, outputting the position of each object in the form of bounding boxes. To pursue more precise localization of objects, SAM-2~\cite{ravi2024sam} is used to get the meticulous segmentation mask of each object. Then, a rule-based algorithm is carried out to judge the spatial relation between objects. Combining visual perception with spatial intelligence, we put them together with the task instruction into GPT-4 and prompt it to decompose the task into subtasks and manage their execution order. The specially designed prompt asks GPT-4 to sequentially perform the following steps: generate a complete execution plan based on the detailed task description and specify which subtask needs to be executed next.


In this way, during the construction of atomic skill library, the VLP agent can effectively decompose end-to-end tasks into multiple subtasks. During practical inference, the VLP agent provides critical low-frequency control signals to plan and guide high-frequency atomic skill execution.

\subsection{VLA Wheel}


For the proposed data-driven approach to skill library generation, any state-of-the-art (SOTA) VLA methods can be leveraged for atomic skill construction. Initially, the VLA model serves as a prompt input to assist the high-level semantic abstraction module in mapping complex subtasks into a structured set of atomic skill definitions. Subsequently, the VLA model facilitates the construction of the skill library through data collection and few-shot learning, enabling efficient and scalable skill acquisition.


The granularity of atomic skills is determined by the performance of VLA models, particularly in terms of their plasticity and adaptability. The plasticity of a VLA model reflects its ability to transition effectively from the pre-trained state to a fine-tuned model, adapting to new robotic platforms. On the other hand, the adaptability of a fine-tuned VLA model demonstrates its capacity to handle diverse objects, scenes, and spatial configurations. Higher levels of plasticity and adaptability lead to a coarser granularity in the definition of atomic skills.

For instance, in the case of the RDT-1B model, we fine-tuned its released pre-trained model on 40 A800 80GB GPUs for atomic skill library construction.
The fine-tuning data comprised 6,000 open-source trajectories and 2,000 proprietary trajectories, collected by a robot utilizing the Mobile ALOHA system design \cite{fu2024mobile} and manufactured by agilex.ai. The fine-tuned VLA model enables the rapid construction of atomic skills. We validated its performance across different sensors, objects, and scenes.

However, we observed that the model's generalization capability for target object positions was limited, and the number of training steps had a significant impact on its behavior. To address these issues, we conducted two experiments: First, we collected a small dataset with varying object positions for a task and performed few-shot training and testing. Specifically, trajectory data was gathered from nine different position points on a table. The results indicated a significant improvement in generalization for object positions, with successful grasping achieved across the entire region defined by these nine points. Second, we conducted a few-shot training step test using 8 L40s GPUs, setting the training steps to 1,000, 2,000, and 4,000. Our findings showed that 1,000 training steps provided the best trade-off between training effectiveness and duration. 

These experiments highlight that the plasticity of pre-trained models, along with the adaptability of fine-tuned models to various objects, scenes, and spatial positions, is critical for skill library construction. These factors serve as inputs to the high-level semantic abstraction module, facilitating the mapping of subtasks into a set of atomic skill definitions. From atomic skill definitions to constructing the skill library, we employed a few-shot fine-tuning approach on the VLA model, leveraging a small number of collected trajectory data for each defined atomic skill. This approach allows for the rapid realization of atomic skills, significantly accelerating the skill library development process.

\begin{figure*}[h]
	\centering
	\includegraphics[width=0.88\textwidth]{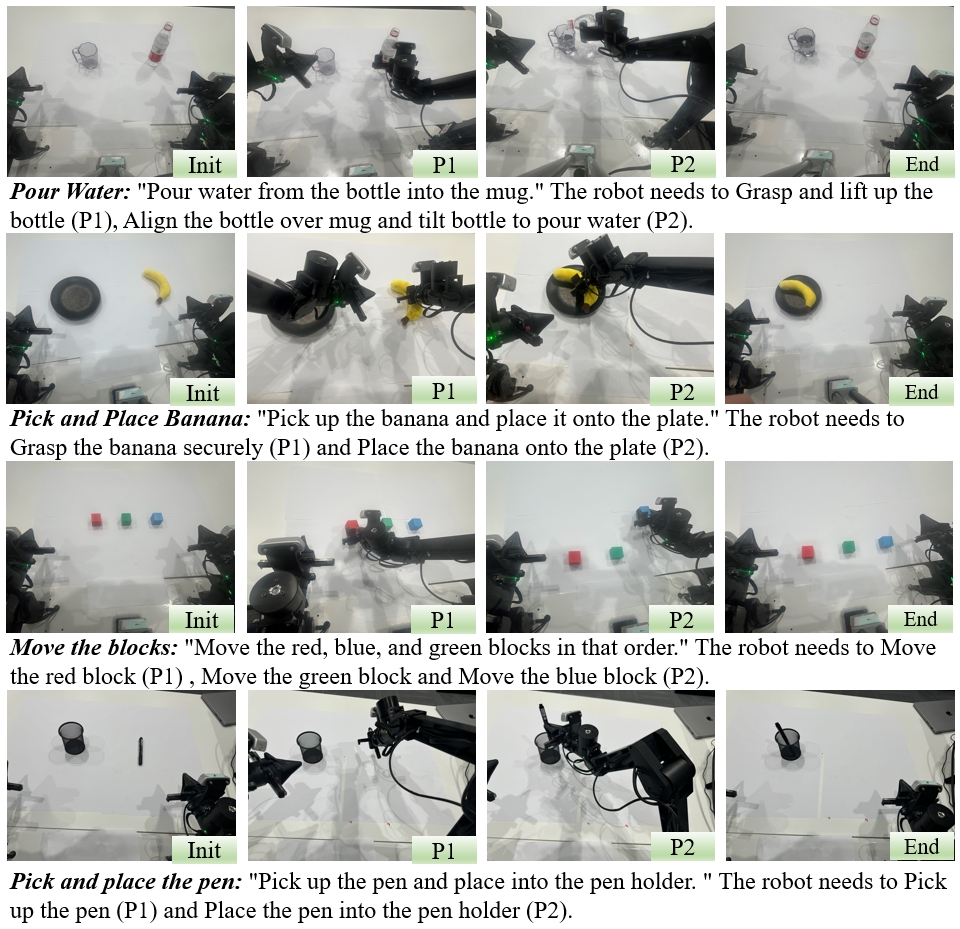}
	\caption{Task definitions and visualization.}
	\label{fig:task_img}	
\end{figure*}

\subsection{Atomic Skill Wheel}
The process of building the atomic skill library is illustrated in Fig.~\ref{fig:framework}. For a set of tasks (A, B, ..., N),  the VLP module decomposes these tasks into corresponding subtasks (\#1, \#2, \#3, \#4, \#5, …, \#a, \#a+1). Next, the high-level semantic abstraction module is used to abstract these subtasks into a set of general atomic skill definitions by a certain granularity, which is determined by the performance of different VLA models. It can be achieved using LLMs like GPT-4. The atomic skill definitions include *1, *2, *3, *4, *5, …, *b, *b+1. Through data collection and fine-tuning of the VLA model, we can ultimately achieve an atomic skill library *1', *2', *3', *4', *5', ..., *b', *b+1'. 



When faced with a new task, TASK N+1, two scenarios can arise. In the first case, the required atomic skills for the new task have already been covered by the existing atomic skill library, our method can directly execute the task without any further adaptation. In the second case, specific atomic skills are missing from the library, the high-level semantic abstraction module is used again. It takes the subtasks of the new task (N+1), the required skill granularity, and the current set of atomic skill definitions as input, producing an updated set of atomic skill definitions. Following this, only additional data collection and fine-tuning for the missing atomic skills are needed, making this approach far more efficient than traditional end-to-end methods. Furthermore, as the atomic skill library dynamically expands, the range of tasks it can handle increases accordingly.

\begin{table*}[t]
\caption{Experimental results compared with other methods. ``ID'' stands for ``In-Distribution," referring to data that falls within the collected data distribution. ``OOD" stands for ``Out-of-Distribution," referring to data that does not fall within the collected data distribution.}
\label{result-table1}
\centering
\begin{tabular}{l|c|c|c|c}
\hline
\multicolumn{5}{c}{Pick up the banana and place it onto the plate : Pick up \textbar \hspace{0.05cm} Place}  \\ 
\cline{1-5}
 & both ID & banana OOD & plate OOD & both OOD \\ 
Octo(End-to-end) & \hspace{0.5cm} 100 \hspace{0.8cm} 80 \hspace{0.5cm} & \hspace{0.5cm} 40 \hspace{0.8cm} 40 \hspace{0.5cm} & \hspace{0.5cm} 80 \hspace{0.8cm} 40 \hspace{0.5cm} & \hspace{0.5cm} 20 \hspace{0.8cm} 20 \hspace{0.5cm} \\ 
Octo(Ours) & \hspace{0.5cm} 100 \hspace{0.8cm} 80 \hspace{0.5cm} & \hspace{0.5cm} 40 \hspace{0.8cm} 40 \hspace{0.5cm} & \hspace{0.5cm} 80 \hspace{0.8cm} 50 \hspace{0.5cm} & \hspace{0.5cm} 40 \hspace{0.8cm} 20 \hspace{0.5cm} \\ 
Octo(Ours-plus) & \hspace{0.6cm} 100 \hspace{0.7cm} 100 \hspace{0.5cm} & \hspace{0.5cm} 60 \hspace{0.8cm} 60 \hspace{0.5cm} & \hspace{0.5cm} 80 \hspace{0.8cm} 60 \hspace{0.5cm} & \hspace{0.5cm} 60 \hspace{0.8cm} 40 \hspace{0.5cm} \\ 
RDT(End-to-end) & \hspace{0.5cm} 90 \hspace{0.9cm} 90 \hspace{0.4cm} & \hspace{0.5cm} 40 \hspace{0.8cm} 40 \hspace{0.5cm} & \hspace{0.5cm} 80 \hspace{0.8cm} 40 \hspace{0.5cm} & \hspace{0.5cm} 60 \hspace{0.8cm} 30 \hspace{0.5cm} \\ 
RDT(Ours) & \hspace{0.5cm} 90 \hspace{0.9cm} 80 \hspace{0.4cm} & \hspace{0.5cm} 50 \hspace{0.8cm} 40 \hspace{0.5cm} & \hspace{0.5cm} 90 \hspace{0.8cm} 70 \hspace{0.5cm} & \hspace{0.5cm} 60 \hspace{0.8cm} 30 \hspace{0.5cm} \\ 
RDT(Ours-plus) & \hspace{0.6cm} 100 \hspace{0.7cm} 100 \hspace{0.5cm} & \hspace{0.5cm} 80 \hspace{0.8cm} 80 \hspace{0.5cm} & \hspace{0.5cm} 100 \hspace{0.8cm} 80 \hspace{0.5cm} & \hspace{0.5cm} 80 \hspace{0.8cm} 70 \hspace{0.5cm} \\  \hline

\multicolumn{5}{c}{Pour water from the bottle into the mug : Grasp \textbar \hspace{0.05cm} Pour}  \\ 
\cline{1-5}
 & both ID & bottle OOD & mug OOD & both OOD \\ 
Octo(End-to-end) & \hspace{0.4cm} 60 \hspace{0.9cm} 0 \hspace{0.5cm} & \hspace{0.4cm} 40 \hspace{0.9cm} 0 \hspace{0.5cm} & \hspace{0.4cm} 60 \hspace{0.9cm} 0 \hspace{0.5cm} & \hspace{0.4cm} 0 \hspace{1.0cm} 0 \hspace{0.4cm} \\ 
Octo(Ours) & \hspace{0.4cm} 60 \hspace{0.9cm} 0 \hspace{0.5cm} & \hspace{0.4cm} 40 \hspace{0.9cm} 0 \hspace{0.5cm} & \hspace{0.4cm} 40 \hspace{0.9cm} 0 \hspace{0.5cm} & \hspace{0.4cm} 0 \hspace{1.0cm} 0 \hspace{0.4cm} \\  
Octo(Ours-plus) & \hspace{0.5cm} 80 \hspace{0.8cm} 30 \hspace{0.5cm} & \hspace{0.5cm} 60 \hspace{0.8cm} 30 \hspace{0.5cm} & \hspace{0.5cm} 80 \hspace{0.8cm} 20 \hspace{0.5cm} & \hspace{0.5cm} 60 \hspace{0.8cm} 20 \hspace{0.5cm} \\ 
RDT(End-to-end) & \hspace{0.5cm} 80 \hspace{0.8cm} 60 \hspace{0.5cm} & \hspace{0.5cm} 60 \hspace{0.8cm} 30 \hspace{0.5cm} & \hspace{0.5cm} 90 \hspace{0.8cm} 40 \hspace{0.5cm} & \hspace{0.5cm} 70 \hspace{0.8cm} 20 \hspace{0.5cm} \\ 
RDT(Ours) & \hspace{0.5cm} 90 \hspace{0.8cm} 60 \hspace{0.5cm} & \hspace{0.5cm} 60 \hspace{0.8cm} 50 \hspace{0.5cm} & \hspace{0.5cm} 90 \hspace{0.8cm} 40 \hspace{0.5cm} & \hspace{0.5cm} 80 \hspace{0.8cm} 40 \hspace{0.5cm} \\ 
RDT(Ours-plus) & \hspace{0.5cm} 90 \hspace{0.8cm} 70 \hspace{0.5cm} & \hspace{0.5cm} 70 \hspace{0.8cm} 60 \hspace{0.5cm} & \hspace{0.5cm} 90 \hspace{0.8cm} 50 \hspace{0.5cm} & \hspace{0.5cm} 90 \hspace{0.8cm} 50 \hspace{0.5cm} \\  \hline

\multicolumn{5}{c}{Pick up the pen and place it into the pen holder : Pick up \textbar \hspace{0.05cm} Place }  \\ 
\cline{1-5}
 & both ID & pen OOD & holder OOD & both OOD \\ 
Octo(End-to-end) & \hspace{0.4cm} 10 \hspace{0.9cm} 0 \hspace{0.5cm} & \hspace{0.4cm} 0 \hspace{1.0cm} 0 \hspace{0.4cm} & \hspace{0.4cm} 10 \hspace{0.9cm} 0 \hspace{0.5cm} & \hspace{0.4cm} 0 \hspace{1.0cm} 0 \hspace{0.4cm} \\ 
Octo(Ours) & \hspace{0.4cm} 10 \hspace{0.9cm} 0 \hspace{0.5cm} & \hspace{0.4cm} 0 \hspace{1.0cm} 0 \hspace{0.4cm} & \hspace{0.4cm} 10 \hspace{0.9cm} 0 \hspace{0.5cm} & \hspace{0.4cm} 0 \hspace{1.0cm} 0 \hspace{0.4cm} \\  
Octo(Ours-plus) & \hspace{0.4cm} 30 \hspace{0.9cm} 0 \hspace{0.5cm} & \hspace{0.4cm} 0 \hspace{1.0cm} 0 \hspace{0.4cm} & \hspace{0.4cm} 20 \hspace{0.9cm} 0 \hspace{0.5cm} & \hspace{0.4cm} 0 \hspace{1.0cm} 0 \hspace{0.4cm} \\  
RDT(End-to-end) & \hspace{0.5cm} 100 \hspace{0.7cm} 70 \hspace{0.6cm} & \hspace{0.5cm} 60 \hspace{0.8cm} 50 \hspace{0.5cm} & \hspace{0.5cm} 100 \hspace{0.7cm} 50 \hspace{0.6cm} & \hspace{0.5cm} 50 \hspace{0.8cm} 30 \hspace{0.5cm} \\ 
RDT(Ours) & \hspace{0.5cm} 100 \hspace{0.7cm} 70 \hspace{0.6cm} & \hspace{0.5cm} 70 \hspace{0.8cm} 50 \hspace{0.5cm} & \hspace{0.5cm} 100 \hspace{0.7cm} 40 \hspace{0.6cm} & \hspace{0.5cm} 70 \hspace{0.8cm} 30 \hspace{0.5cm} \\ 
RDT(Ours-plus) & \hspace{0.5cm} 100 \hspace{0.7cm} 90 \hspace{0.6cm} & \hspace{0.5cm} 100 \hspace{0.7cm} 70 \hspace{0.6cm} & \hspace{0.5cm} 100 \hspace{0.7cm} 70 \hspace{0.6cm} & \hspace{0.5cm} 80 \hspace{0.8cm} 40 \hspace{0.5cm} \\  \hline

\end{tabular}
\end{table*}

\section{Experiment}
We incorporate our framework with various well-performing VLA models to experimentally compare it with traditional end-to-end methods, aiming to answer the following questions:
\begin{itemize}
    \item When collecting trajectory data under the same physical setting, can our method achieve the comparable performance of the end-to-end method with less data?
    \item By collecting the same amount of data, can our method achieve better performance than the end-to-end method?
    \item When dealing with new tasks, can our method still work well with few or without additional data?
    \item Is our method applicable to different backbone VLA models while maintaining effectiveness and efficiency?
\end{itemize}

\subsection{Experimental Setup}
\paragraph{Baseline.} Our method can be integrated into different end-to-end VLA models. We use the RDT-1B and Octo model as baselines and conduct experiments on the ALOHA dual-arm robot. According to previous research and our test, other widely used VLAs, like OpenVLA, are hard to adapt to the dual-arm Mobile ALOHA hardware settings. Thus, we exclude these models from our experiment.

\paragraph{Tasks.} 

We selected four challenging tasks to evaluate the performance of our methods from different dimensions, including complex scenarios that the model may encounter in real-world tasks, such as various positions of objects and sophisticated operations. The first three tasks were specifically designed to validate the data efficiency and performance of our method, while the fourth task was used to assess its adaptability to new tasks. Detailed task definitions and visualizations are provided in Figure \ref{fig:task_img}.

\begin{table*}[t]
\caption{Experimental results comparing the block-grasping task with other methods}
\label{result-table2}
\centering
\begin{tabular}{l|c|c|c|c}
\hline
\multicolumn{5}{c}{Move the red, blue, and green blocks in that order. }  \\ 
\cline{1-5}
 & Red-Green-Blue & Red-Blue-Green & Blue-Green-Red & Green-Red-Blue \\ 
Octo(End-to-End) & \hspace{0.2cm} 80 \hspace{0.6cm} 80 \hspace{0.6cm} 60 \hspace{0.2cm} & \hspace{0.2cm} 0 \hspace{0.6cm} 0 \hspace{0.6cm} 0 \hspace{0.2cm} & \hspace{0.2cm} 0 \hspace{0.6cm} 0 \hspace{0.6cm} 0 \hspace{0.2cm} & \hspace{0.2cm} 0 \hspace{0.6cm} 0 \hspace{0.6cm} 0 \hspace{0.2cm}  \\ 
Octo(Ours) & \hspace{0.2cm} 80 \hspace{0.6cm} 60 \hspace{0.6cm} 60 \hspace{0.2cm} & \hspace{0.2cm} 80 \hspace{0.45cm} 50 \hspace{0.45cm} 60 \hspace{0.2cm} & \hspace{0.2cm} 60 \hspace{0.45cm} 70 \hspace{0.45cm} 50 \hspace{0.2cm} & \hspace{0.2cm} 60 \hspace{0.45cm} 60 \hspace{0.45cm} 60 \hspace{0.2cm}  \\ 
RDT(End-to-End) & \hspace{0.1cm} 100 \hspace{0.55cm} 90 \hspace{0.6cm} 70 \hspace{0.2cm} & \hspace{0.2cm} 0 \hspace{0.6cm} 0 \hspace{0.6cm} 0 \hspace{0.2cm} & \hspace{0.2cm} 0 \hspace{0.6cm} 0 \hspace{0.6cm} 0 \hspace{0.2cm} & \hspace{0.2cm} 0 \hspace{0.6cm} 0 \hspace{0.6cm} 0 \hspace{0.2cm}  \\ 
RDT(Ours) & \hspace{0.1cm} 100 \hspace{0.55cm} 80 \hspace{0.6cm} 70 \hspace{0.2cm} & \hspace{0.05cm} 100 \hspace{0.45cm} 80 \hspace{0.5cm} 90 \hspace{0.2cm} & \hspace{0.05cm} 100 \hspace{0.45cm} 90 \hspace{0.5cm} 90 \hspace{0.2cm} & \hspace{0.05cm} 100 \hspace{0.45cm} 90 \hspace{0.5cm} 80 \hspace{0.2cm}  \\  \hline

\end{tabular}
\end{table*}

\paragraph{Data.} 
We collected trajectory data samples for fine-tuning the backbone VLA model. The number of demos for each task is as follows:
\begin{itemize}
    \item Pour Water: 3 different bottle positions and 3 different mug positions, 3 demos for each setting, leading to 27 demos for the end-to-end fine-tuning, 9 and 9 demos for the skill-based VLA fine-tuning.
    \item Pick \& Place Banana: 4 different banana positions and 2 different plate positions, 3 demos for each setting, leading to 24 demos for end-to-end VLA fine-tuning, 9 and 6 demos for skill-based VLA fine-tuning.
    \item Pick \& Place Pen: 3 different bottle positions and 3 different mug positions, 3 demos for each different settings, leading to 27 demos for the end-to-end VLA fine-tuning, 9 and 9 demos for the skill-based VLA fine-tuning.
    \item Move blocks in order: 10 demos of the Red-Green-Blue order for the end-to-end VLA fine-tuning, 10 demos of moving red, green, and blue block respectively for the skill-based VLA fine-tuning.
\end{itemize}

\paragraph{Metric.} For a specific task, we evaluate performance by comparing the task success rates under different settings. The success rate is the number of successful trials divided by the total number of trials. To ensure fairness, we conduct 10 trials for each method across the four tasks. Below, we use the task of taking a banana and placing it on a plate as an example. First, both the banana and the plate are placed at positions within the training set for 10 trials. Next, the banana is placed at a position outside the training set while the plate remains within it for another 10 trials. Then, the banana is placed within the training set while the plate is placed outside it for 10 trials. Finally, both the banana and the plate are placed at positions outside the training set for 10 trials.

\subsection{Results Analysis}
We identified four complex tasks and conducted comparative experiments using our method on both the Octo and RDT-1B models. The results are shown in Table \ref{result-table1} and Table \ref{result-table2}. It should be noted that ``End-to-end" refers to the traditional end-to-end method, ``Ours" refers to maintaining the same distribution of data points as in end-to-end data collection but with a smaller data volume, and ``Ours-plus" refers to maintaining the same data volume as in end-to-end data collection, resulting in a larger distribution of data points. 

Here, we conduct a detailed analysis of the experimental results to address the three questions raised earlier.

Q1: From Table \ref{result-table1}, it can be observed that our method integrated with Octo or RDT-1B shows comparable performance to corresponding end-to-end methods. All the success rates of our method are no less than the end-to-end method. For instance, in the task of picking up a bottle and pouring water, when both the bottle and the cup are positioned outside the training set, our method achieves a 20\% improvement in success rate. This demonstrates that our approach requires less data under the same distribution while achieving comparable or even better performance, effectively addressing ``data explosion."

Q2: Similarly, Table \ref{result-table1} shows when fine-tuned on the same amount of demonstrations, our method significantly improves the success rate regardless of Octo or RDT-1B VLA. For example, in the task of placing a banana onto a plate, when both the banana and the plate are positioned outside the training set, our method increases the success rate by 40\%. This improvement stems from gathering data from more diverse locations while maintaining the same data volume, thereby enhancing the model's generalization capabilities.

Q3: Table \ref{result-table2} shows the comparison of performance on new tasks between our method and end-to-end method. We can find that end-to-end methods are limited to the known task and can not handle new tasks at all,  while our method can effectively perform required atomic skills across different new tasks for successful execution. 

Q4: Table~\ref{result-table1} shows improvement of data efficiency and manipulation performance of our method regardless of Octo or RDT-1B compared to their corresponding end-to-end method. Table~\ref{result-table2} shows our method can better adapt to new tasks regardless of the backbone VLA. These results indicate that our method can be effectively applied to various VLA models, improving data efficiency, manipulation performance, and adaptability to new tasks.

\section{Conclusion}

In this work, we propose a data-driven framework for atomic skill library construction, termed the ``Three-wheeled Self-Driven Atomic Skill Library Construction Method", to address the ``data explosion'' issue caused by the traditional end-to-end embodied manipulation strategy. Overall, our approach automatically defines and updates a set of atomic skills in a data-driven manner and then realizes these skills through data collection and VLA fine-tuning. Extensive experiments on real-world scenarios show the data efficiency and generality of our method. We hope this work can inspire future research on significant solutions to ``data explosion''.

\section{Acknowledgments}
We would like to acknowledge d-robotics.cc, Realman robotics and Agilex robotics for their support in providing the experimental sensors and robotic platforms. Additionally, we acknowledge the authors of RDT-1B for their significant contributions to the development of the VLA model. This work was supported by the National Key Research and Development Program of China (NO.2024YFE0211000).

\bibliographystyle{named}
\bibliography{ijcai25}

\end{document}